%% file: main.tex
\def\year{2015}
\RequirePackage{snapshot}
\documentclass[letterpaper]{article}
\usepackage{aaai}
\usepackage{times}
\usepackage{helvet}
\usepackage{courier}
\usepackage{color}
\usepackage{booktabs}
\usepackage{graphicx}
\usepackage{verbatim}
\usepackage{url}
\usepackage{amsmath}
\usepackage{caption}
\usepackage{subcaption}

\newcommand{\mycite}[1]{\citeauthor{#1}~\shortcite{#1}}
\renewcommand{\vec}[1]{\mathbf{#1}}
\newcommand{\example}[1]{\emph{#1}}

\begin{document}
%

\title{Emoticons vs. Emojis on Twitter: A Causal Inference Approach}
\author{Umashanthi Pavalanathan \and Jacob Eisenstein \\
  School of Interactive Computing \\
  Georgia Institute of Technology \\
  Atlanta, GA 30308 \\
  {\tt \{umashanthi + jacobe\}@gatech.edu}}
\maketitle

\input{abstract}
\input{intro}

\input{data}
\input{method}

\input{results}
\input{related}
\input{discussion}

\begin{small}
\bibliographystyle{aaai} 
\bibliography{cite-strings-short,cites,cite-definitions}
\end{small}
\end{document}

%% file: abstract.tex
\begin{abstract}
Online writing lacks the non-verbal cues present in face-to-face communication, which provide additional contextual information about the utterance, such as the speaker's intention or affective state. 
To fill this void, a number of orthographic features, such as emoticons, expressive lengthening, and non-standard punctuation, have become popular in social media services including Twitter and Instagram. Recently, emojis have been introduced to social media, and are increasingly popular. This raises the question of whether these predefined pictographic characters will come to replace earlier orthographic methods of paralinguistic communication. In this abstract, we attempt to shed light on this question, using a matching approach from causal inference to test whether the adoption of emojis causes individual users to employ fewer emoticons in their text on Twitter.
\end{abstract}

%% file: intro.tex
\section{Introduction}
People are changing writing to express themselves in online settings, often through the use of non-standard orthographies, such as emoticons (e.g., \example{(:}) and letter repetitions (e.g., \textit{coooolll})~\cite{dresner2010functions,kalman2014repetition}. The introduction of emojis is a potentially dramatic shift in online writing, potentially replacing these user-defined linguistic affordances with predefined graphical icons. With the ability to access a large number of colorful and expressive emoji pictographs, will users stop employing non-standard orthographies for expressive communication in social media?

In this abstract, we address the question of whether the individual users' adoption of emojis reduces the frequency of emoticons used in their tweets. From a sample of mostly English tweets, we extracted authors who were early adopters of emojis, and consider them as the treatment group. To measure the causal effect of emoji adoption on emoticon usage, we choose another set of authors (control) who were not yet using emojis at the same time as the treatment group, and compare the differences in emoticon usage of these two groups between a period of an year. We matched each author in the treatment group with an author in the control group, based on their emoticon usage rate before the treatment period. If the individuals in the treatment group reduce their emoticon usage more than the individuals in the control group, this would suggest that emojis are competing with emoticons, and may eventually reduce the amount of non-standard orthography in social media.

\begin{figure}
\includegraphics[width=.45\textwidth]{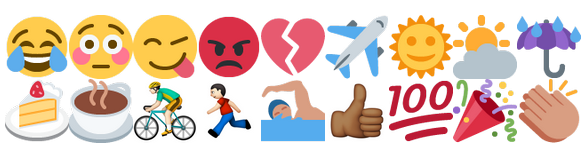}
\vspace{-6pt}
\caption{Examples of emoji characters used in Twitter (created using http://www.iemoji.com)}
\vspace{-10pt}
\label{fig:emoji_eg}
\end{figure}

\begin{figure}
\centering
\includegraphics[width=.35\textwidth]{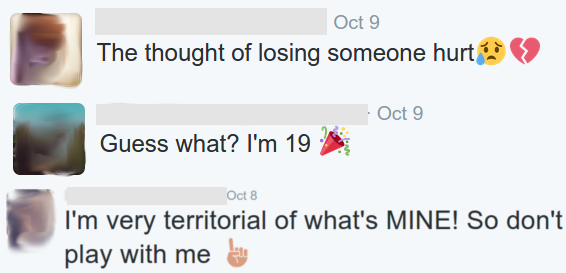}
\vspace{-6pt}
\caption{Examples tweets using emoji characters}
\vspace{-10pt}
\label{fig:emoji_eg_tweets}
\end{figure}

Emojis are ``picture characters'' that originated for mobile phones in Japan in the late 1990s, but recently became popular worldwide in text messaging and social media with the adoption of smartphones supporting input and rendering of emoji characters. In contrast to emoticons, which are created from ASCII character sequences, emojis are represented by unicode characters, and are continuously increasing in number with the introduction of new characters in each new unicode version.\footnote{\url{http://www.unicode.org/reports/tr51/index.html#Selection_Factors}} 
Emoji characters include not only faces, but also concepts and ideas such as weather, vehicles and buildings, food and drink, or activities such as running and dancing (Figure~\ref{fig:emoji_eg}, and example tweets in Figure~\ref{fig:emoji_eg_tweets}). 
Emoji Tracker reports real time emoji use on Twitter.\footnote{\url{http://www.emojitracker.com/}} 

In computer mediated communication (CMC), emoticons are interpreted as ``emotion icons'', primarily as a way to represent facial expressions, such as smile, in the absence of non-verbal cues~\cite{walther2001impacts}. However, later research has shown that emoticons are not just representation of affective stances; they play many other roles in written communication such as showing author intention, sociocultural differences, and author identity~\cite{derks2007emoticons,schnoebelen2012you,park2013emoticon}. In particular, ~\mycite{dresner2010functions} situate the usage of emoticons in CMC between the extremes of non-language and language.

We hypothesize that individuals who adopt emojis tend to use fewer emoticons, indicating that emojis are replacing this particular form of orthographic paralinguistic communication. 
We use a matching approach to causal inference to test our hypothesis using observational data from Twitter. Next, we describe the dataset, our study design and report results. Then we first briefly discuss related work
and conclude with discussion and future work.

%% file: data.tex
\section{Dataset}
We gathered a corpus of tweets from February 2014 to August 2015, using Twitter's streaming API. We removed retweets (repetitions of previously posted messages) by excluding messages which contain the ``retweeted\_status'' metadata or the ``RT'' token. We included only authors who have written at least five tweets on average each month and removed authors who have written more than 10\% of their tweets in any language other than English. 

\subsection{Extracting Emoji and Emoticon Tokens}
\begin{figure}
\centering
(a) \includegraphics[width=.48\textwidth]{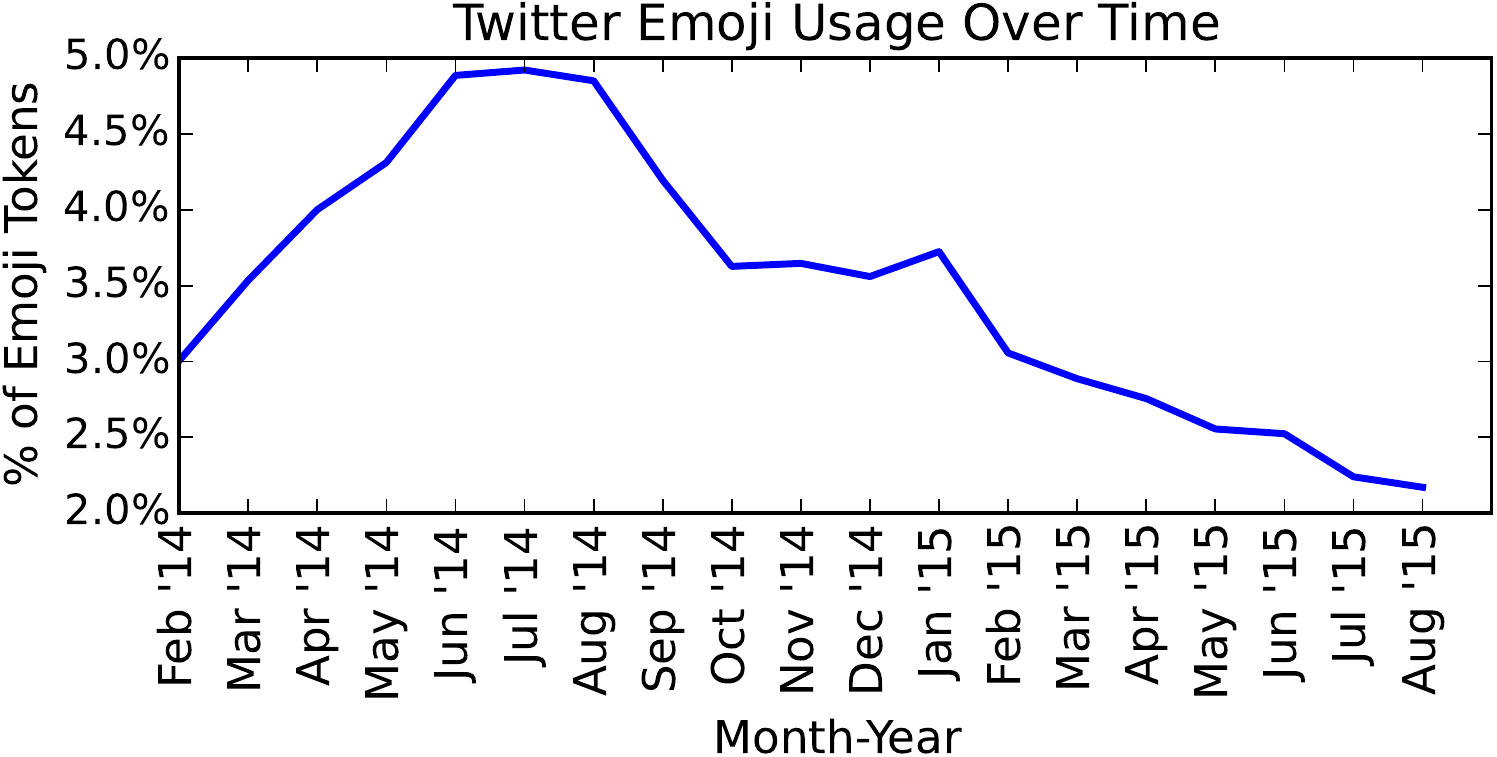}
\\[1pt]
(b) \includegraphics[width=.48\textwidth]{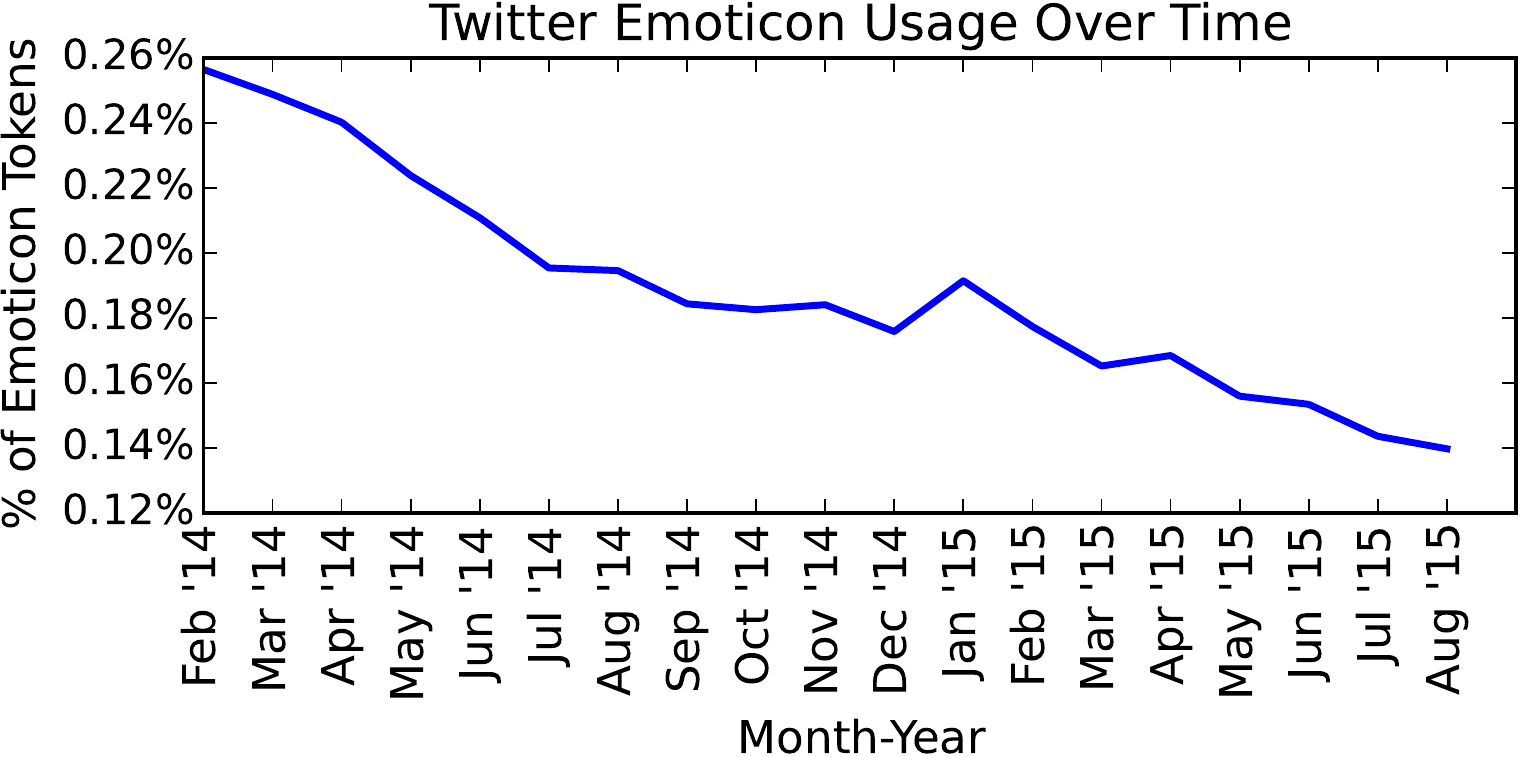}
\vspace{-10pt}
\caption{Temporal trend of percentage of (a) Emoji, and (b) Emoticon tokens in tweets from our sample.}
\label{fig:token_rate}
\end{figure}

To extract emoji characters from tweets, we converted the messages into unicode representation and used regular expressions to extract unicode characters in the ranges of the ``Emoji \& Pictographs'' category of unicode symbols (other categories include non-Roman characters such as different numbering systems and mathematical symbols). Using this method we identified 1,235 unique emoji characters in a random sample of tweets spanning a period of more than an year (February 2014 to August 2015). Figure~\ref{fig:token_rate}a shows the percentage of emoji character tokens (i.e., $\frac{\text{\# of emoji tokens}}{\text{\# of total tokens}} \times 100\%$) over time in our our sample of mostly English tweets.~\footnote{Note that although there is a decreasing trend in emoji usage rate after a peak in June-August 2014 in this sample, emoji usage rate shows an upward trend in a sample of unfiltered tweets, indicating an increasing popularity of emojis on Twitter.}

As there is no comprehensive list of Twitter emoticons (and new emoticons get introduced over time), we used a data-driven approach to identify emoticons. We constructed regular expressions (e.g., two or more characters with at least one non-alpha numeric character, not containing money/percent/time symbols, etc.) to retrieve an initial set of emoticon-like tokens, and then manually annotated all the items that made up 95\% cumulative frequency of emoticon-like tokens, looking at their usage on random examples of tweets. After removing tokens that are not used as emoticons, there were 44 and 52 unique emoticons extracted from tweets of March 2014 and March 2015, respectively. In both cases, the twenty most frequent emoticons made up 90\% of all emoticon tokens. Figure~\ref{fig:token_rate}b shows the percentage of emoticon symbols (i.e., $\frac{\text{\# of emoticon tokens}}{\text{\# of total tokens}} \times 100\%$) over time in our sample. 
Table~\ref{tab:basic-stats} shows the basic usage statistics of emojis and emoticons for the same sample.

\begin{table}[t!]
\centering
  \begin{small}
    {
		\begin{tabular}{lrr} \toprule
		\textbf{Measure}& \textbf{Emoticon}&\textbf{Emoji} \\
		\midrule
		\% of messages with feature & 2.27\% & 13.69\%\\
		\% of authors who have used &&\\
		  the feature at least once & 50.65\% & 	63.11\%\\[6pt]
		\multicolumn{3}{l}{\textbf{Of the messages with emoticons/emojis, percent with...}} \\
		\hspace{1em}feature at the beginning &0.72\% & 13.57\%\\
		\hspace{1em}feature at the end  &46.15\%& 50.55\%\\
		\hspace{1em}feature as sole token &0.15\% &0.30\%\\
		\hspace{1em}tokens of only this feature type &0.16\%&0.92\%\\
		\bottomrule
		\end{tabular}
	}		
	\vspace{-6pt}
    \caption{Emoticon and emoji usage statistics}
    \vspace{-10pt}
    \label{tab:basic-stats}
\end{small}
\end{table}

%% file: method.tex
\section{Study Design}
The causal inference question is whether the introduction of emojis causes orthographic features such as emoticons to decrease in frequency. In a randomized experiment, the treatment group would be individuals who used emojis, and the control group would be individuals who did not. We would choose these individuals to have similar pre-treatment characteristics: zero usage of emojis, and comparable usage of emoticons. The treatment effect is the difference in the usage of emoticons after the treatment.

\subsection{Treatment and Control Groups}
We approximate this randomized experimental setup using observational data. We consider the month of March 2014 as the pre-treatment period and the month of March 2015 as the post-treatment period, and our primary analysis is based on tweets from these two months. Before placing tweet authors into treatment and control groups, we first selected a pool of authors with following criteria: users who have written at least five tweets in each month, and who have written less than 10\% of their tweets in any language other than English. This criteria was chosen because extraction of non-standard words and emoticon symbols depends on the language of the message.

For the treatment group, we selected authors who had not used any emoji characters in March 2014, and who used at least five emoji characters in March 2015. Authors who had not used any emoji characters in both March 2014 and March 2015 were chosen into the control group. We matched authors in the treatment and control group based on their emoticon usage rate (i.e., $\frac{\text{\# of emoticon tokens}}{\text{\# of total tokens}}$, matching up to two decimal points) prior to the treatment. We extracted 5,205 such treatment-control author pairs from our treatment and control groups and used these authors for our primary analysis. Figure~\ref{fig:emoji_rate} shows the distribution of emoji usage rates in both the treatment and control groups, before and after treatment.

\subsection{Causal Inference Framework}
When computing the treatment effect, we need to account for the control group's outcome as well. This is because the overall emoticon usage rate is changing with time as shown in Figure~\ref{fig:token_rate}b, and hence there might be changes in the emoticon usage rates by the control group even without any intervention. The treatment effect is computed in terms of the following quantities:\\[1pt]
\begin{small}
\begin{tabular}{llll}
&&$Y_{i}^{t}$& the difference between the post-treatment and\\
&&&  pre-treatment emoticon usage rates for author $i$,\\
&&&  who is in the treatment group $t$\\
&&$Y_{j}^{c}$ &the difference between the post-treatment and\\
&&&  pre-treatment emoticon usage rates for author $j$,\\
&&&  who is in the control group $c$\\
&&$\bar{Y}^{t}$ & the average difference between post-treatment \\
&&& and pre-treatment emoticon usage rates for the \\
&&& treatment group: $\bar{Y}^{t} = \frac{1}{n_{t}} \sum_{i \in T} Y_{i}^{t}$\\
&&$\bar{Y}^{c}$ & the average difference between post-treatment \\
&&& and pre-treatment emoticon usage rates for the \\
&&& control group: $\bar{Y}^{c} = \frac{1}{n_{c}} \sum_{j \in C} Y_{j}^{c}$\\
\\[-6pt]
\end{tabular}
\end{small}

\noindent We can then define the average treatment effect as,\\[-12pt]
\begin{align}
ATE = \bar{Y}^{t} - \bar{Y}^{c}
\end{align}

%% file: results.tex
\section{Results}

\begin{figure}
\includegraphics[width=.46\textwidth]{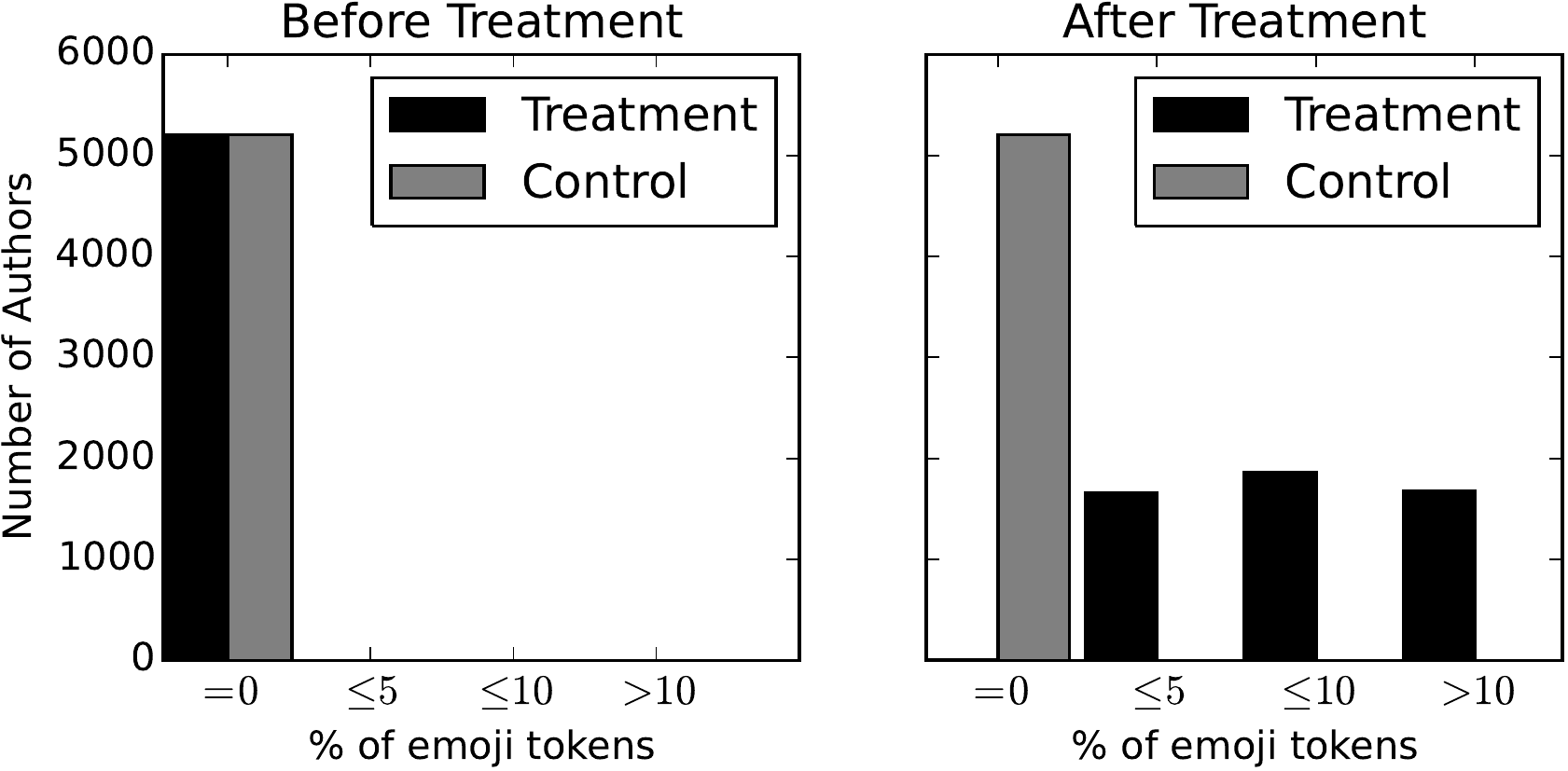}
\vspace{-6pt}
\caption{Emoji usage of treatment and control groups}
\vspace{-10pt}
\label{fig:emoji_rate}
\end{figure}

\begin{figure}
\includegraphics[width=.46\textwidth]{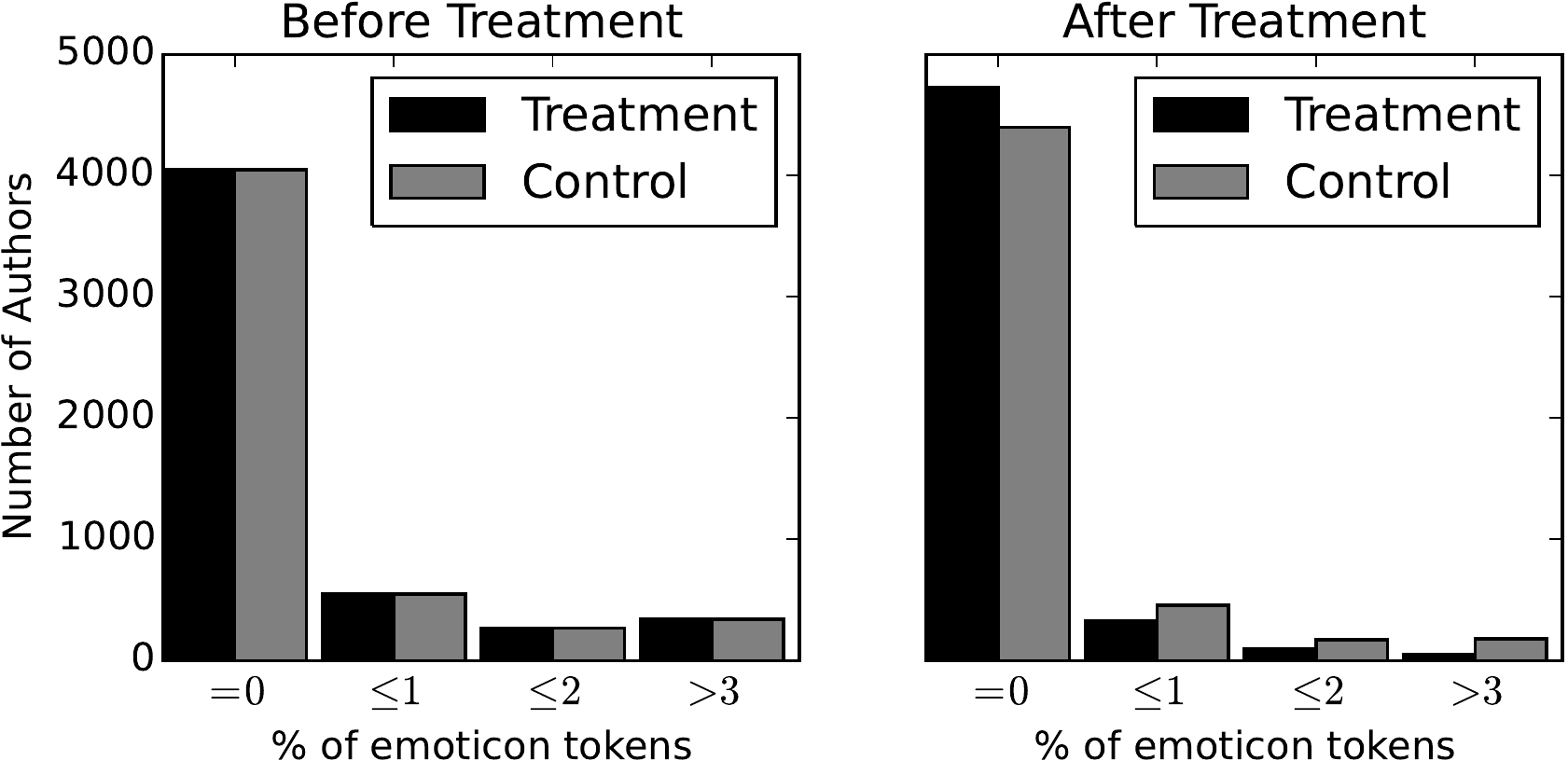}
\vspace{-6pt}
\caption{Emoticon usage of treatment and control groups}
\vspace{-10pt}
\label{fig:emoticon_rate}
\end{figure}

The distributions of authors' emoticon usage before and after treatment are shown in Figure~\ref{fig:emoticon_rate}. By design, both the treatment and control groups have similar distributions of emoticon usage before treatment, with an average emoticon usage of 0.50\% (i.e., 0.50\% of the tokens from tweets of both the treatment and control users are emoticons). We see a decrease in emoticon usage for the treatment group after treatment, as in the distribution shown in Figure~\ref{fig:emoticon_rate}. After the treatment intervention, the treatment group has an average emoticon token usage of 0.14\%, while the control group has an average of 0.30\% of emoticon tokens. The overall emoticon usage trend shown in Figure~\ref{fig:token_rate}b supports the decreased emoticon usage in the control group, even without any intervention. The difference between the emoticon usage of the treatment and control groups is statistically significant by a paired t-test ($ t=-11.11, p\approx 1 \times 10^{-27}$), with an average treatment effect of 0.17\% decrease in emoticon symbols per token. A preliminary analysis of the rate of change for different emoticon types shows that happy and playful emoticons, such as :-) and :P, had higher rate of decrease than sad emoticons such as :( . This may be because of seemingly fewer sad face emojis compared to emojis with happy and playful faces; further analysis is left for future work.

%% file: related.tex
\section{Related Work}
\subsection{Causal Inference and Social Media Analysis}
Although causal inference framework has been widely used in fields such as epidemiology and political science  to design and analyze randomized experiments, this approach was rare in social media research until recently~\cite{muchnik2013social,king2014reverse}.
Observational studies of causal phenomena are susceptible to confounds because subjects are not randomly assigned to treatment and control groups as in randomized experiments. Several statistical techniques are developed to mitigate these confounds in observational studies of causal phenomena including matching~\cite{rosenbaum1983central,ho2007matching} and stratification~\cite{frangakis2002principal}. There has been some recent work employing causal inference techniques such as matching in large scale quantitative studies using observational social media data~\cite{culotta2015matchedsamples,cheng2015antisocial}. We apply these approaches to the analysis of linguistic style for the first time.



\subsection{Linguistic Functions of Emoticons}
Emoticons in textual communication has been long studied and considered to be the expressions of emotions that mimic nonverbal cues in speech~\cite{rezabek1998visual,wolf2000emotional,crystal2006language}. However, there is an increasing understanding that the meaning of emoticons in CMC goes beyond affective stances and varies with social context and author identity~\cite{derks2007emoticons,schnoebelen2012you,park2013emoticon}. Specifically, ~\mycite{dresner2010functions} draw concepts from speech act theory~\cite{austin1975things} and argue that emoticons are indications of the speaker's intentions, the illocutionary force of the accompanied textual utterances. They identify three broad linguistic functions of emoticons: (1) as emotion indicators, mapped directly onto facial expressions (e.g. happy or sad), (2) as non-emotional meaning, mapped conventionally onto facial expressions (e.g., joking), and (3) as an indication of illocutionary force that do not map conventionally onto facial expressions (e.g., a smiley mitigating the investment in an utterance). Emojis seem to be able to play similar roles (see below), which is why we contrast them with emoticons in this study.

\subsection{Emojis in Textual Communication}
With the increased popularity of emojis in textual communication, researchers have started to explore the roles of emojis in textual communication. 
~\mycite{kelly2015emojiappropriation} interviewed a culturally diverse set of 20 participants about the differential use of emojis in mediated textual communication with close personal ties and found that beyond expressing emotions, emojis are used for other purposes such as maintaining a conversational connection, permitting a playful interaction, and creating a shared and secret uniqueness within a particular relationship. 
~\mycite{novak2015emojisentiment} developed a sentiment lexicon for emojis using their usage in tweets.\footnote{\url{http://kt.ijs.si/data/Emoji_sentiment_ranking/}} Our study is the first to consider how emojis compete with emoticons to communicate paralinguistic content.

%% file: discussion.tex
\section{Discussion and Future Work}
Our results show that the Twitter users who adopt emojis tend to reduce their usage of emoticons, in comparison with matched users who do not adopt emojis. These results support our hypothesis that the emojis compete with emoticons, and that the introduction of emojis can lead to a decline in orthographic variation.
Of course, since Twitter has a restriction on the number of characters, in some sense all linguistic features compete for linguistics functions. Nonetheless, the overwhelming majority of Twitter messages are not near the character limit~\cite{eisenstein2013bad}, indicating that this is unlikely to be the main reason for the decrease in emoticon characters --- rather, it seems more likely that emojis are replacing emoticons in fulfilling the same paralinguistic functions. The next steps in this study are to look at how emojis compete with non-standard language such as expressive lengthening (e.g., \textit{cooooolll!!!} ), non-standard words (e.g., \textit{gud}), and abbreviations (e.g., \textit{lol}).

%